\documentclass{article}
\usepackage{spconf}
\usepackage{amssymb}
\usepackage{graphicx}
\usepackage{graphicx,color}
\usepackage{amssymb, array}
\usepackage[margin=0.750in]{geometry}

\usepackage[cmex10]{amsmath}

\usepackage{algorithmic}

\usepackage{array}

\usepackage{mdwmath}
\usepackage{mdwtab}

\usepackage{tikz}
\usepackage{epigraph}

\usepackage{eqparbox}

\usepackage{pgffor}%
\usepackage{pgfplots}%
\pgfplotsset{compat=1.5}
\usepackage{forloop}%
\usepackage{tikz}
\usepackage[font=footnotesize]{subfig}

\usepackage{stfloats}

\usepackage{url}

\newcommand{\rad}{10}
\newcommand{\ns}{3}
\newcommand{\nc}{36}
\newcommand{\fov}{60}
\newcommand{\orproj}%
{   
	\coordinate (a) at (-\rad,-\rad);
	\coordinate (b) at (-\rad,\rad);
	\coordinate (c) at (\rad,\rad);
	\coordinate (d) at (\rad,-\rad);
	\draw (a)--(b);
	\draw (b)--(c);
	\draw[blue,very thick] (c)--(d);
	\draw (d)--(a);
	\foreach \x in {1,...,\ns}
	{
		\pgfmathsetmacro\z{\rad-2*\rad*(\x-1)/(\ns)};
		\coordinate (w) at (-\rad,\z);
		\coordinate (u) at (\rad,\z);
		\draw (w)--(u);
	}
}

\DeclareMathOperator{\arctan2}{arctan2}
\DeclareMathOperator{\sign}{sign}

\newcommand{\norproj}%
{   
	  \pgfmathsetmacro\an{\fov/2};
	  \pgfmathsetmacro\tana{tan(\an)};
	  \pgfmathsetmacro\xw{\rad/\tana};
	  \pgfmathsetmacro\tmpa{max(abs(\rad),abs(\xw))};
	  \pgfmathsetmacro\tmpb{min(abs(\rad),abs(\xw))};
	  \pgfmathsetmacro\tmpc{\tmpb/\tmpa};
	  \pgfmathsetmacro\foc{sqrt(1+\tmpc*\tmpc)*\tmpa-\rad};
		\pgfmathsetmacro\ha{\foc*\tana};
		\pgfmathsetmacro\hb{(2*\rad+\foc)*\tana};
	\coordinate (a) at (-\rad,-\hb);
	\coordinate (b) at (-\rad,\hb);
	\coordinate (c) at (\rad,\ha);
	\coordinate (d) at (\rad,-\ha);
	\draw (a)--(b);
	\draw (b)--(c);
	\draw[blue,very thick] (c)--(d);
	\draw (d)--(a);
	\foreach \x in {1,...,\ns}
	{
		\pgfmathsetmacro\zw{\hb-2*\hb*(\x-1)/(\ns)};
		\pgfmathsetmacro\zu{\ha-2*\ha*(\x-1)/(\ns)};
		\coordinate (w) at (-\rad,\zw);
		\coordinate (u) at (\rad,\zu);
		\draw[black] (w)--(u);
	}
}

\def\x{{\mathbf x}}

\graphicspath{{images/}}
\title{On the Accuracy of Point Localisation in a Circular Camera-Array}
\name{Alireza Ghasemi \hspace*{20pt} Adam Scholefield\hspace*{20pt} Martin Vetterli \thanks{This work was supported by the Commission for Technology and Innovation (CTI) project no. 14842.1 PFES-ES.}}
\address{\vspace*{-0pt}
School of Computer and Communication Sciences\\
\'{E}cole Polytechnique F\'{e}d\'{e}rale de Lausanne\\
	}
\small
\begin{document}
%
\maketitle

\begin{abstract}

Although many advances have been made in light-field and camera-array image processing, there is still a lack of thorough analysis of the localisation accuracy of different multi-camera systems. By considering the problem from a frame-quantisation perspective, we are able to quantify the point localisation error of circular camera configurations. Specifically, we obtain closed form expressions bounding the localisation error in terms of the parameters describing the acquisition setup. 

These theoretical results are independent of the localisation algorithm and thus provide fundamental limits on performance. Furthermore, the new frame-quantisation perspective is general enough to be extended to more complex camera configurations.



\end{abstract}
\begin{keywords}
Image sampling, Image sequences, Reconstruction algorithms, Signal quantization. 
\end{keywords}
%

\section{Introduction}
A key advantage of multi-camera systems is their ability to recover the 3-D information lost during projection. However, a formal, quantitive analysis on the accuracy of such recovered information is still lacking. This is surprising given the long-standing interest in these systems \cite{DBLP:journals/pami/LuhK85,trivedi1988can,aghajan2009multi} and improvements that they can make to vision algorithms \cite{farshidi2005active,porikli2003multi,ghasemi2014scalee,ghasemi2014icassp,ghasemi2014lcav}.


Recently, a range of novel camera-array architectures, such as the Lytro light-field camera \cite{georgiev2013lytroo} and PiCam mobile camera array architecture \cite{venkataraman2013picam} have been introduced, altering the direction of modern consumer digital photography. In the presence of such developments a rigorous treatment of this topic is needed.

In the two-view (stereo-imaging) case, probabilistic error analyses have been carried out \cite{blostein1987error,rodriguez1990stochastic}; however in the case of more than two views, no such analysis exists. Excellent multi-view scene reconstruction algorithms have been proposed \cite{kolmogorov2002multi,wilburn2005high}, but they have not considered a quantitative error analysis. Chai et al \cite{chai:plenopticsampling} studied the frequency spectrum of the plenoptic function \cite{adelson1991plenopticc}, which has been used to find the critical sampling rate for error-free image-based rendering of different classes of scenes \cite{zhang:spectralIBR,do2012bandwidth,gilliam2013spectrum}. Raynor et. al. \cite{raynor2013plenoptic} analysed the error in range finding using a plenoptic camera utilising a model based on Gaussian noise.


In this paper, we study circular camera-arrays, in an algorithm-independent way, through different error criteria. We relate the problem to frame quantisation \cite{goyal1998quantized,beferull2003efficient,rangan2001recursive,thao1994reduction}, which is well studied in the information theory community and has a rigorous mathematical foundation. In particular, Cvetkovi\'c \cite{cvetkovic1999source} has analysed the partitions induced by uniform scalar quantisation of expansions of $\mathbb{R}^2$.

We show that these results can be utilised to bound the localisation error of circular camera systems under orthogonal projections. Furthermore, we extend these results to the perspective projection case.

More precisely, we show that, when the number of cameras is sufficiently large, the localisation error is upper bounded by a term that decreases quadratically with respect to the number of cameras. This shows that the error tends to zero as the number of cameras tends to infinity. These results hold for the majority of the region of interest, however there is an exception, which we fully quantify. Finally, we provide numerical simulations, which verify the quadratic bound.



\section{Problem Setup}
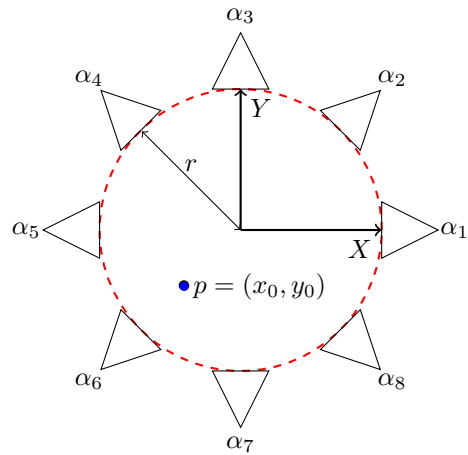
\begin{figure}[t!]
	\begin{center}
		\begin{tikzpicture}[scale=0.75]
		\draw[->,thick] (0,0) -- node[below left,pos=1] {$X$}(2.5,0);
		\draw[->,thick] (0,0) -- node[below right,pos=1] {$Y$}(0,2.5);
		\draw[<->] (-1.75,1.75) -- node[above] {$r$}(0,0);
		\node at (-1,-1) (P1) {$\bullet$};
		\draw[blue,fill=blue] (P1) circle (2pt);
		\draw[dashed, red, thick] (0,0) circle (2.5);
		\node[right] at (P1) {$p=(x_0,y_0)$};
		
		\foreach \i in {1,2,3,4,5,6,7,8}
		{
			\pgfmathsetmacro\angl{(\i+1)*45};
			\begin{scope}[rotate=\angl]
				\draw[] (0,-3.5)--(0.5,-2.5)--(-0.5,-2.5) -- cycle;
				\node[rotate=0] at (0,-3.8) {$\alpha_{\i}$};
			\end{scope}
		}
		\end{tikzpicture}
	\end{center}
	\vspace*{-20pt}
	\caption{A point $p=(x_0,y_0)$ in the unit-circle is captured by $m=8$ cameras uniformly spaced around a circle. The problem is to localise this point from these $m$ projections.
	}
	\label{setup}
\end{figure}

We are interested in the error in reconstructing a 3-D point from different multi-camera configurations. For simplicity, we will limit our analysis to circular-camera setups, such as the configuration shown in Fig. \ref{setup}. Furthermore, we will assume that the world is 2-D, rather than 3-D, resulting in 1-D images. This allows us to easily visualise the results and the extension to 3-D is relatively straight forward.

More formally, we wish to reconstruct a point $p=(x_0,y_0)$, located within a circle of radius $r$, from $m$ images, captured with $m$ cameras positioned uniformly on the perimeter of the circle and oriented towards the origin. We will refer to the interior of this circle as the region of interest. Let us index the cameras from $1$ to $m$ and associate each one with its angular location $\alpha_i$, measured anti-clockwise from the $X$-axis. Figure \ref{setup} depicts this configuration for the case of eight cameras.
\begin{figure}[t!]
	\begin{center}
	\tikzstyle{block} = [draw,fill=blue!20,minimum size=2em]
	\tikzstyle{matrix} = [draw,fill=green!20,minimum size=5em]
	\def\radius{.7mm} 
	\tikzstyle{branch}=[fill,shape=circle,minimum size=3pt,inner sep=0pt]
	\usetikzlibrary{arrows}
		\begin{tikzpicture}[>=latex']

		\node at (2,-6) (block6) {};

		\node at (0,-6) (F) {$p$};
		\node[] at (7.5,-6) (s) {$\mathbf{s}_{i}$};			
				\foreach \y in {6} {
					\node at (0,-\y) (input\y) {};
					\node at (0.5,-\y) (inp\y) {};
					\node[block] at (1,-\y) (block\y) {$\mathcal{T}_i$};
					\draw[->] (block\y.east) -- node[above] {$q_i$} +(1.0,0);
				}
				
				\foreach \y in {6} {
					
					\draw[->] (F) -- (block\y.west);
					
				}
			\foreach \y in {6} {
				\node[block] at (2.70,-\y) (psf\y) {$\Psi$};
				\draw[->] (psf\y.east) -- node[above] {\hspace{5pt}$h_i(u)$} +(1.5,0);
			}
			
			\foreach \y in {6} {
				\node[block] at (4.9,-\y) (sk\y) {$\Xi$};
				\draw[color=black,-] (sk\y.east) --  node[above] {} +(0.5,0);	
				\draw[color=black,-o] (5.74,-6) --  node[above] {} +(1.0,0.5);	
				\draw[color=black,o->] (6.74,-6) --  node[above] {} +(0.6,0.0);	
				\draw[->] (6.74,-6) to[bend right] (6.74,-5.5);
			}

		\tikzstyle{s}=[shift={(0mm,\radius)}]

		\end{tikzpicture}
	\end{center}
	\vspace*{-10pt}
	\caption{Camera acquisition model.}
	\label{camera_acquisition}
\end{figure}
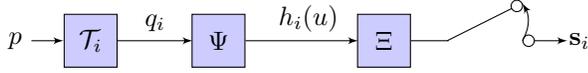

Our goal is to find an estimator, $\hat{p}=(\hat{x_0},\hat{y_0})$, for the location, $p=(x_0,y_0)$, and analyse its behaviour and error bounds. To do this we need to understand how the point $p$ is captured in each camera. To this end, we utilise the camera acquisition model depicted in Fig. \ref{camera_acquisition}. 

Here, $\mathcal{T}_i$ is a projection operator, which maps the 2-D point $p$ onto the 1-D image plane of the $i$-th sensor. The 1-D projection is then subjected to a convolution with a point-spread function (PSF), $\Psi$, which models the blurring of the camera lens. Finally, the continuous-domain to discrete-domain sampling is achieved by a convolution with the integration kernel, $\Xi$, before ideal sampling. The details and justification of each of these steps is given in the following subsections.

\subsection{The Projection Operator}
\label{projop}
The first and most critical stage of the acquisition model is the projection of the point $p$ onto the image plane of the $i$-th sensor. This image plane is the line tangent to the circle at the point $[\cos\alpha_i, \sin\alpha_i]^T$.

We will consider both orthogonal and non-orthogonal projections, as depicted in Fig. \ref{fig:projections}.

The orthogonal projection case has been widely studied and lays the foundations of computed tomography and magnetic resonance imaging. 
In this case, the projection point, $q^o_i$, is given by the following simple inner product:
\begin{equation}
	\begin{split}
		q^o_i=\left\langle p,\phi(\alpha_i)\right\rangle=-x_0\sin\alpha_i+y_0\cos\alpha_i,
	\end{split}
\end{equation}
where $\phi(\alpha_i)=\begin{bmatrix}-\sin\alpha_i& \cos\alpha_i\end{bmatrix}^T$ is the unit vector parallel to the $i$-th image plane that points anti-clockwise around the circle.

The non-orthogonal, perspective, projection is a more accurate model of traditional cameras and is equivalent to the pin-hole camera model. In this case, the projections of $p$ onto the sensors are no longer inner products. In fact, the location of the projected point, $q^{no}_i$, is given by
\begin{equation}
	\begin{split}
		q^{no}_i=\frac{f}{r+f-\langle p,{\phi_i}^\perp \rangle}\left\langle p,\phi_i \right\rangle,
	\end{split}
\end{equation}
where $f$ is the focal length of the camera, $\phi_i=\phi(\alpha_i)$ and $\phi_i^\perp=\phi^\perp(\alpha_i)=\begin{bmatrix}\cos\alpha_i&\sin\alpha_i\end{bmatrix}^T$ is the unit vector, orthogonal to $\phi_i$, that points towards the centre of the $i$-th image plane. Note that, as $f\rightarrow \infty$, $q^{no}_i\rightarrow q^o$. 

Finally, we will assume that the cameras have equal length image planes, denoted $\tau_s$, chosen so that each camera covers the whole region of interest. I.e. $\tau_s=2r$ in the orthogonal case and $\tau_s=\frac{2fr}{\sqrt{f^2+2fr}}$ in the non-orthogonal case.

%
%
%
%
%
%
%

\tikzset{
  pics/carc/.style args={#1:#2:#3}{
    code={
      \draw[pic actions] (#1:#3) arc(#1:#2:#3);
    }
  }
}

\begin{figure}[t!]
\captionsetup[subfloat]{position=top,labelformat=empty}
	\centering
	\vfil
	\subfloat[]{
				\begin{tikzpicture}[scale=0.8]
		
				\node at (3.05,3) (P2) {};	
				\draw[fill=blue] (P2) circle (3pt);
				\node[right] at (P2) {\hspace{3pt}};



				\node at (1.5,-1.0) (C1) {};
				\draw[above,very thick] (-1.05,1)--(3.95,1) ;
				
				\draw[<->] (-1.0,-1.2)--node[below] {$\tau_s$} (4.0,-1.2) ;
				\draw[<->] (4.1,-1) -- node[right] {$f$} (4.1,1) ;
				\draw[fill=black]  (C1) circle (2pt);
				\draw[above,dashed,-] (1.5,3)-- node[right] {} (1.5,1) ;
				
				\draw [red,thick,dashed,domain=205:335] plot ({1.5+3*cos(\x)}, {4+3*sin(\x)});

				\draw[<->] (1.5,4)-- node[left] {$r$} (-1,2.3) ;
				
				\draw[above,very  thick,->] (1.5,4)-- (2.5,4) node[right] {$\phi_i$};
				\draw[above,very  thick,->] (1.5,4)-- (1.5,3) node[above left] {$\phi_i^\perp$} ;

				\draw[above,very thick,->] (1.5,4)--(P2) node[above] {$p$} ;
				\draw[fill=black] (1.5,1) circle (1pt) ;				

				
				\draw[dashed] (P2) -- (C1) ;
				

		
				\node at (2.28,1.0) (Q2) {};
				\draw[fill=blue] (Q2) circle (1pt) node[below]{$q^{no}_i$};
				\node[above] at (Q2) {\hspace{20pt}};

				\node at (3.05,1.0) (Q2o) {};
				\draw[fill=blue] (Q2o) circle (1pt) node[below]{$q^{o}_i$};
				\node[above] at (Q2o) {\hspace{20pt}};
				\draw[dashed] (P2) -- (Q2o) ;
		

						
				\end{tikzpicture}
		\label{nop}
		}
	\caption{The orthogonal ($q^{o}_i$) and non-orthogonal ($q^{no}_i$) projection operators applied to a 2-D point $p$, within a circle of radius $r$.}
	\label{fig:projections}
\end{figure}
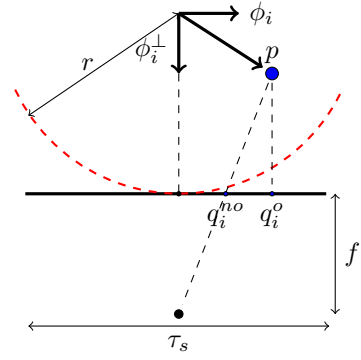

\subsection{The Image Acquisition Pipeline}

The pin-hole camera model is equivalent to a perspective projection; however, a more realistic model, such as the one depicted in Fig. \ref{camera_acquisition}, must also account for the camera optics and sampling process. 

Let us represent the projection of the point $p$, onto the image plane of the $i$-th sensor, as the 1-D function $g_i(u)=\delta\left(u-q_i\right)$; i.e., a Dirac at the location $q_i$. The effect of the camera lens can then be modelled by the following convolution:
\begin{equation}
	\begin{split}
		h_i(u)=g_i(u) \ast \Psi(u),
	\end{split}
\end{equation}
where $\Psi(u)$ is the point-spread function \cite{westheimer1962light} which is commonly modeled by the Airy disc.
%
%
%
%


To model the sampling operation of an imaging sensor we apply a further convolution followed by ideal sampling. For an $n$-pixel sensor, we choose the sampling kernel, $\Xi(u)$, to be a box function with support 
$w=\frac{\tau_s}{n}$; i.e., the pixel width. This models the fact that image sensors usually work by integrating all the light rays which fall into each pixel region \cite{russ2010image}. Finally, the ideal sampling has period $w$.

\section{Point localisation as a frame quantisation problem}
\begin{figure*}[b]
		\hrulefill
		\normalsize
		\newcounter{MYtempeqncnt}
		\setcounter{MYtempeqncnt}{\value{equation}}
		\setcounter{equation}{10}
		\begin{equation}
	\begin{split}
		\delta=\frac{1}{\sin ( \alpha_{j_1} -  \alpha_{j_2})}
		\begin{pmatrix}
			\cos \alpha_{j_2} &   -\cos \alpha_{j_1}\\
			\sin \alpha_{j_2}  &  -\sin \alpha_{j_1}
		\end{pmatrix}
		\begin{pmatrix}
			\sin \alpha_{j_1}-\sin \alpha_{j_1}^\prime  &  \cos \alpha_{j_1}^\prime-\cos \alpha_{j_1}\\
			\sin \alpha_{j_2}-\sin \alpha_{j_2}^\prime  &  \cos \alpha_{j_2}^\prime-\cos \alpha_{j_2}
		\end{pmatrix}
		\begin{bmatrix}
			x_0\\y_0
		\end{bmatrix}. \label{eq:loc_err_expanded}
	\end{split}
\end{equation}
		
		\setcounter{equation}{\value{MYtempeqncnt}}
\end{figure*}

In the previous section, we reviewed a traditional image acquisition model which is applicable to a wide range of imaging devices, including traditional cameras and computation tomography (CT) devices.

We will now show how this can be interpreted as a quantised frame expansion, which will allow us to derive closed-form expressions for the worst-case localisation error.

To see that this interpretation is valid, note that the camera vectors, $\phi(\alpha_i)$, form a frame in $\mathbb{R}^2$, which we will denote $\Phi$. Therefore, we can consider the projections of the point $p$ onto all the image planes as $p$ projected onto the frame $\Phi$; i.e., $q=\Phi p$.

In addition, if we assume the point activates only a single pixel in each camera\footnote{If the PSF causes the point to be spread across multiple pixels, one could exploit more advanced sampling techniques that achieve sub-pixel accuracy.}, the acquisition model, depicted in Fig. \ref{camera_acquisition}, is equivalent to the following quantisation:
\begin{equation}
	\begin{split}
		s_i=Q_w\left(q_i\right),
	\end{split}
\end{equation}
where $Q_w$ is the quantisation function defined as\footnote{This definition is valid when $n$ is even. For odd $n$, $Q_w(y)=\sign(y)\left\lfloor \frac{|y|}{w} +\frac{1}{2}\right\rfloor w$.}
\begin{equation}
	\begin{split}
		Q_w(y)=\left\lfloor \frac{y}{w} \right\rfloor w + \frac{w}{2}.
	\end{split}
\end{equation}
In quantisation terminology, the pixel width, $w$, is the quantisation error.

\subsection{Orthogonal Projection}

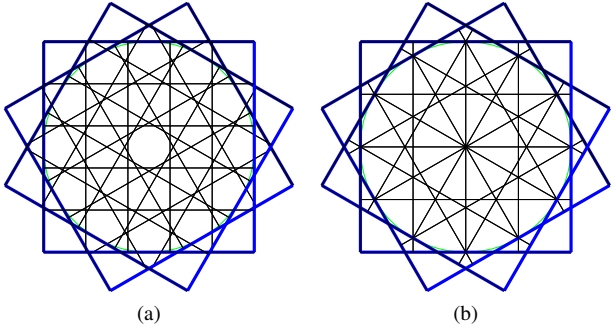
\begin{figure}[!t]
	\vspace{-2mm}
	\centering
	\subfloat[]{
	\begin{tikzpicture}[scale = 0.14]
			\renewcommand{\rad}{10}
			\renewcommand{\ns}{5}
			\renewcommand{\nc}{12}
			\renewcommand{\fov}{25}
			\draw[color=green] (0,0) circle (\rad);	
			\foreach \x in {1,...,\nc}
			{
					\pgfmathsetmacro\angl{\x*360/\nc};
					\begin{scope}[rotate=\angl]
						\orproj
					\end{scope}
			}
			\end{tikzpicture}%
		\label{t1}
		}
		\hfil
		\subfloat[]{
		\begin{tikzpicture}[scale = 0.14]
				\renewcommand{\rad}{10}
				\renewcommand{\ns}{4}
				\renewcommand{\nc}{12}
				\renewcommand{\fov}{25}
				\draw[color=green] (0,0) circle (\rad);	
				\foreach \x in {1,...,\nc}
				{
						\pgfmathsetmacro\angl{\x*360/\nc};
						\begin{scope}[rotate=\angl]
							\orproj
						\end{scope}
				}
				\end{tikzpicture}%
			\label{t1}
			}
	\vspace{-1mm}
	\caption{Partitionings of the circle of interest for orthogonal projection with an odd (left) and even (right) number of samples. After sampling, it is impossible to distinguish between two points that lie in the same region.}
	\label{epiplane}
	\vspace{-10pt}
\end{figure}

We will first investigate if we can localise the point $p$ with infinite precision by using an infinite number of cameras. In the orthogonal projection case, it was shown, in \cite{cvetkovic1999source}, that this is possible iff $\|p\|_2\geq w$.

This condition can be intuitively understood from Fig. \ref{epiplane}: as we add more cameras, there is a central circle which, in the case of odd $n$, is not further divided and, in the case of even $n$, is only divided through the origin. 

To understand this more precisely, let us consider the projection of the point $p$ onto an image plane at angle $\alpha$ as a function of $\alpha$:
\begin{equation}
	\begin{split}
		q^o(\alpha)=-x_0\sin\alpha+y_0\cos\alpha=\|p\|_2\sin(\alpha+\beta),
	\end{split}
\end{equation}
where $\beta=\arctan2(y_0,-x_0)$.
For even $n$, the quantised version, 
\begin{equation}
	\begin{split}
		s_i=Q_w\left(q^o(\alpha)\right)=\left\lfloor \frac{\|p\|_2\sin(\alpha+\beta)}{w} \right\rfloor w + \frac{w}{2},
	\end{split}
\end{equation} 
has discontinuities at angles $\alpha_{d_i}$, where $\|p\|_2\sin(\alpha_{d_i}+\beta)=k_i w$ for some integer $k_i$. When $\|p\|_2<w$, this only occurs at one threshold ($k_i=0$), which occurs at the two angles $\alpha_{d_1}=-\beta$ and $\alpha_{d_2}=\pi-\beta$. As $m\rightarrow \infty$, we are guaranteed to have cameras at these locations; however, since the angles are $\pi$ radians apart, the image planes are parallel and we can only localise $p$ to be in the central circle of radius $w$ on the line connecting the two camera centres, i.e. on the line with a quantisation threshold of zero.

When $\|p\|_2\geq w$, there are more than two discontinuity angles and, since $p\in\mathbb{R}^2$, it can be perfectly reconstructed from its projections onto two non-parallel image planes:
\begin{equation}
	p=\left\langle p,\phi(\alpha_{j_1}) \right\rangle \tilde{\phi}(\alpha_{j_1})+\left\langle p,\phi(\alpha_{j_2}) \right\rangle \tilde{\phi}(\alpha_{j_2}),
\end{equation}
where $\alpha_{j_1}$ and $\alpha_{j_2}$ are the angles of the image planes, $\tilde{\phi}(\alpha_{j_1})=\frac{1}{\sin(\alpha_{j_2}-\alpha_{j_1})}\begin{bmatrix}\cos\alpha_{j_2} & \sin\alpha_{j_2}\end{bmatrix}^T$ is the dual of $\phi(\alpha_{j_1})$ and $\tilde{\phi}(\alpha_{j_2})$ is the dual of $\phi(\alpha_{j_2})$. 

Of course, we can only approximate $p$ because we only have access to quantised versions of these projections:
\begin{equation*}
	\hat{p}=Q_w\left(\left\langle p,\phi(\alpha_{j_1}) \right\rangle\right) \tilde{\phi}(\alpha_{j_1})+Q_w\left(\left\langle p,\phi(\alpha_{j_2}) \right\rangle\right) \tilde{\phi}(\alpha_{j_2}).
\end{equation*}
To investigate the accuracy of this approximation, consider two imaginary cameras placed at unknown angles $\alpha_{j_1}^\prime$ and $\alpha_{j_2}^\prime$ such that $Q_w\left(\left\langle p,\phi(\alpha_{j_i}) \right\rangle\right)=\left\langle p,\phi(\alpha_{j_i}^\prime) \right\rangle$, $i=1,2$; i.e., $\alpha_{j_i}^\prime$  is slightly shifted from $\alpha_{j_i}$, $i=1,2$, so that the projection of $p$ is exactly between the two quantisation boundaries.

Using these cameras, we can write the approximated point, $\hat{p}$, as
\begin{align}
	\hat{p}&=\left\langle p,\phi(\alpha_{j_1}^\prime) \right\rangle \tilde{\phi}(\alpha_{j_1})+\left\langle p,\phi(\alpha_{j_2}^\prime) \right\rangle\tilde{\phi}(\alpha_{j_2})\nonumber\\
	&=\left(\tilde{\phi}(\alpha_{j_1})\phi(\alpha_{j_1}^\prime)^T+\tilde{\phi}(\alpha_{j_2})\phi(\alpha_{j_2}^\prime)^T\right)p.
\end{align}
The localisation error, $\delta=p-\hat{p}$, is thus given by
\begin{equation}
	\delta=\left(I-\tilde{\phi}(\alpha_{j_1})\phi(\alpha_{j_1}^\prime)^T-\tilde{\phi}(\alpha_{j_2})\phi(\alpha_{j_2}^\prime)^T\right)p.\label{eq:loc_err}
\end{equation}

Substituting the expressions for the image plane vectors and their duals into \eqref{eq:loc_err} and applying basic algebraic manipulations yields \eqref{eq:loc_err_expanded}.\addtocounter{equation}{1}

By applying $\left|\sin a-\sin b \right|\leq \left| a-b \right|$ and $\left| \alpha_{j_i}-\alpha_{j_i}^\prime \right| \leq \frac{2\pi}{m}$ to \eqref{eq:loc_err_expanded}, we can bound the localisation error as
\begin{equation}
	\|\delta\|_2^2\leq \frac{4\pi^2(x_0+y_0)^2}{m^2}\leq \frac{8\pi^2r^2}{m^2}.\label{eq:orthogonal_quadratic_bound}
\end{equation}

This means $\|p-\hat{p}\|^2_2$ is upper bounded by a term that decreases quadratically with respect to $m$.

\subsection{Non-Orthogonal Projection}\label{nortcomp}
\begin{figure}[t]
	\centering
	\begin{tikzpicture}[scale = 0.15]
		\renewcommand{\rad}{8}
		\renewcommand{\ns}{5}
		\renewcommand{\nc}{5}
		\renewcommand{\fov}{30}
		\draw[color=green] (0,0) circle (\rad);	
		\foreach \x in {1,...,\nc}
		{
				\pgfmathsetmacro\angl{\x*360/\nc};
				\begin{scope}[rotate=\angl]
					\norproj
				\end{scope}
		}
		\end{tikzpicture}	
		\label{sdatasvdd}
	
	\caption{Partitioning of the circle of interest for non-orthogonal projection with an odd number of samples. After sampling, it is impossible to distinguish between two points that lie in the same region.}
	\label{cr}
\end{figure}
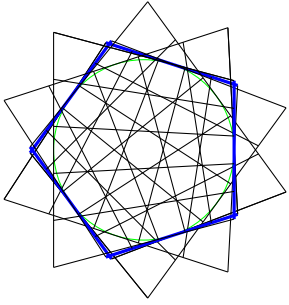

Figure \ref{cr} shows an example partition for non-orthogonal, perspective, projection with an odd number of pixels. Visually, we can see a similar central circle, which is not subdivided by adding further cameras. In the orthogonal case, the radius of this circle was equal to the pixel width $w$. In the non-orthogonal case, the radius increases to 	
$\left(1+\frac{r}{f}\right)w$,
 which approaches $w$ as $f\rightarrow\infty$.

To bound the approximation error, $\delta=\|\hat{p}-p\|_2$, we again analyse the partitionings of quantised frame expansions. In the orthogonal case, the partitioning is created from intersecting rectangles whose shorter edges have length $w$. In the non-orthogonal case, we have trapezoids, instead of rectangles, with sides of length $w$ and $\left(1+2\frac{r}{f}\right)w$.

It follows that each trapezoid has a greater area than the corresponding rectangle with side length $w$. Therefore, the following bound holds in the non-orthogonal case:
\begin{equation}
	\begin{split}
		\| \delta \|_2^2 \leq 8\pi^2\frac{r^2}{m^2}.
	\end{split}
\end{equation}

\section{Simulation Results}
In the previous section, we showed how a circular-camera array divides the region of interest into a finite number of partitions. Examples of such partitionings are shown in Figs. \ref{epiplane} and \ref{cr}\footnote{You can generate your own partitionings using the web app located at {\tt http://rr.epfl.ch/demos/multicam}.}.

%

The interesting semantic behind the constituent regions is that, after projection, we can not distinguish between two points that fall into the same region, meaning there is an inverse relationship between the number of partitions in the circle of interest and the mean squared localisation error. 

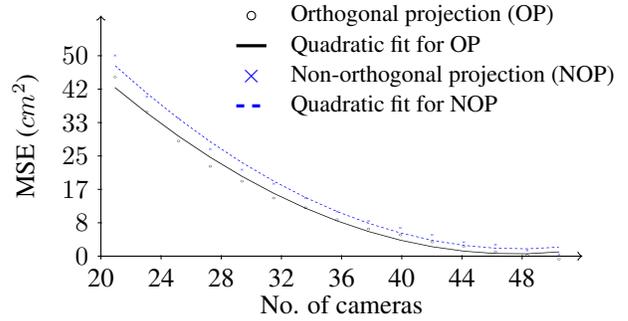
\begin{figure}[t!]
	\begin{center}
		\begin{tikzpicture}[scale=0.4]
		\node at (7.8,3.25) (P1) {\includegraphics[width=2.4in,height=1.1in]{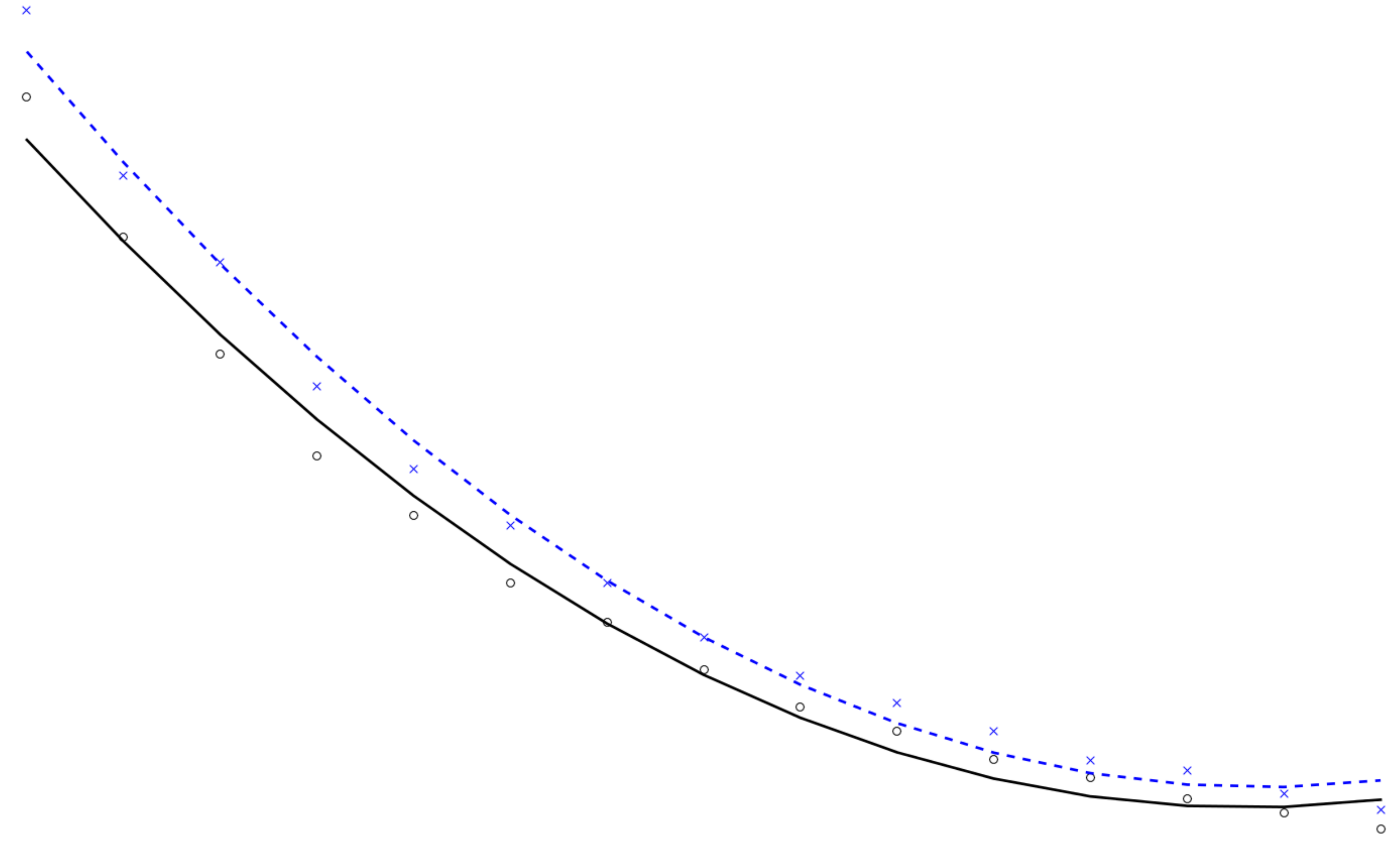}};
%

		\draw[black] (5,8) circle (4pt);
		\draw[very thick,black] (4.5,7) -- (5.5,7) ;
		\node[blue] at (5,6) {$\times$};
		\draw[very thick,dashed,blue] (4.5,5) -- (5.5,5) ;

		\node[right] at (6,8) {\small Orthogonal projection (OP)};
		\node[right] at (6,7) {\small Quadratic fit for OP};
		\node[right] at (6,6) {\small Non-orthogonal projection (NOP)};
		\node[right] at (6,5) {\small Quadratic fit for NOP};

		\draw[->] (0,0) -- coordinate (x axis mid) (16,0);
		\draw[->] (0,0) -- coordinate (y axis mid) (0,8);
		    	\foreach \x in {20,24,...,50}
		    	{
		    			\pgfmathsetmacro\pos{(\x-20)*15/30};
		    			\draw (\pos,1pt) -- (\pos,-3pt) node[anchor=north] {\x};
		    				    	
		    	}
		     	\foreach \y in {0,100,...,600}
		    	{
				    	\pgfmathsetmacro\pos{\y/90};
				    	\pgfmathsetmacro\err{\y/12};
		    			\draw[] (1pt,\pos) -- (-3pt,\pos) node[anchor=east] {\pgfmathprintnumber[fixed,precision=0]{\err}
		    			}; 
		    	}
		     	      
			\node[below=0.4cm] at (x axis mid) {No. of cameras};
			\node[rotate=90, above=0.7cm] at (y axis mid) {MSE ($cm^2$)};
		\end{tikzpicture}
	\end{center}
	\vspace*{-20pt}
	\caption{Mean squared error for up to $50$ three-sample cameras, in both the orthogonal and non-orthogonal cases.
	}
	\label{growth}
\end{figure}

Figure \ref{growth} depicts the mean squared localisation error for a circular-camera array with three-sample sensors and a varying number of cameras. For non-orthogonal projection, a focal length of $f=r$ was used, which leads to a realistic field of view for pinhole cameras. The error values have been well approximated by the reciprocal of a polynomial of degree $2$. This suggests that the error decreases quadratically as the number of cameras increases. 

Moreover, in Fig. \ref{growth}, the error decreases slightly faster for orthogonal projection, which corresponds to the theory of Section \ref{nortcomp}. Although this shows a benefit of orthogonal projection, we should note that the minimum sensor size required to cover the region of interest is much smaller when using perspective projection, as we computed in Section \ref{projop}. 

\section{Conclusion}

We have reformulated the problem of localising a 3-D point from its images, in a circular camera array, as a frame quantisation problem. This reformulation allows us to derive closed-form worst-case bounds for the localisation error.

We showed that the localisation error can be made arbitrarily low by increasing only the number of cameras. Moreover, we extended the results for orthogonal projections to non-orthogonal projections, which are more common in camera architectures. 
 
In our reformulation, we assumed that the point activates only a single pixel in each camera. We believe that this assumption could be relaxed, by adjusting the quantisation noise model appropriately.

Furthermore, we believe that the frame quantisation interpretation could be used to derive similar results for other multi-camera setups.

\bibliographystyle{IEEEbib}
\bibliography{icip2015}
\end{document}